\documentclass{article}
\usepackage[utf8]{inputenc}
\usepackage{amsfonts}
\usepackage{spconf,amsmath,graphicx}
\usepackage{xcolor}
\graphicspath{{./figures/}}
\usepackage{subfigure}
\usepackage[T1]{fontenc}
\usepackage{csquotes}
\usepackage{float}
\usepackage{bm}

\newtheorem{thm}{Theorem}[section]

\newtheorem{prop}[thm]{Proposition}

\name{Aniket Pramanik, Mathews Jacob\thanks{This work is supported by NIH R01AG067078.}}

\address{The University of Iowa, Iowa City, USA}
\begin{document}

\title{Accelerated parallel MRI using memory efficient and robust monotone operator learning (MOL)}

\maketitle

\begin{abstract}
Model-based deep learning methods that combine imaging physics with learned regularization priors have been emerging as powerful tools for parallel MRI acceleration. The main focus of this paper is to determine the utility of the monotone operator learning (MOL) framework in the parallel MRI setting. The MOL algorithm alternates between a gradient descent step using a monotone convolutional neural network (CNN) and a conjugate gradient algorithm to encourage data consistency. The benefits of this approach include similar guarantees as compressive sensing algorithms including uniqueness, convergence, and stability, while being significantly more memory efficient than unrolled methods. We validate the proposed scheme by comparing it with different unrolled algorithms in the context of accelerated parallel MRI for static and dynamic settings.
\end{abstract}

\section{Introduction}

Compressive sensing (CS) algorithms have revolutionized accelerated Magnetic Resonance Imaging (MRI). These methods consider the recovery of images from noisy and highly undersampled multi-channel measurements by posing the recovery as a convex optimization problem. The benefits of CS  algorithms include theoretical guarantees on uniqueness,  convergence, and stability. In recent years, model-based deep-learning (MoDL) has gained tremendous success in parallel MRI \cite{hammernik2018learning,aggarwal2018modl}. These algorithms alternate between a gradient descent step to minimize data consistency, followed by a denoising step using a CNN. 
The alternating optimization blocks are unrolled for few iterations and are trained end-to-end with weights shared across iterations. The image quality of the reconstructions offered by deep unrolled (DU) approaches often surpasses the ones from CS algorithms, in addition to being significantly more efficient in computation during inference. 

%

The central challenge with DUs, which involves algorithm unrolling, is its high memory demand during training. This drawback restricts its utility in high-dimensional applications including 2D+time and 3D applications. Several strategies that trade computational complexity for reduce memory footprint have been introduced to reduce the memory demands \cite{chen2016training,kellman2020memory}. The deep equilibrium (DEQ) framework was introduced as a memory-efficient alternative where non-expansive optimization blocks are iterated until convergence to a fixed point \cite{gilton2021deep}. Since DEQ relies on fixed point iterations rather than back-propagation, it reduces memory demand by a factor of the number of iterations, while its computational complexity is comparable to an unrolled network \cite{bai2019deep}; the tradeoffs offered by this scheme are better than the above discussed strategies \cite{chen2016training,kellman2020memory}. Unlike unrolled methods,the backpropagation steps in DEQ will be accurate only if the forward iterations converge. The convergence of several DEQ schemes were rigorously analyzed in \cite{gilton2021deep,ryu2019plug}. However, these theoretical convergence guarantees are unfortunately not valid in the highly undersampled setting, when the forward model is low-rank or ill-conditioned.

We had recently introduced a DEQ scheme to address the limitations of DUs in \cite{pramanik2022improved}.  This algorithm inherits many of the desirable properties of CS methods including uniqueness of the solution, guaranteed convergence, and robustness to input perturbations. The approach relies on a forward-backward splitting algorithm, where the score function rather than its proximal map, is modeled by a CNN; the algorithm alternates between a gradient step to improve the prior probability and a conjugate gradient algorithm to enforce data consistency using a damping parameter $\alpha$.  We constrain  the score network to be a monotone operator, which we show, is a necessary and sufficient condition for the fixed point of the iterations to be unique. Current algorithms such as MoDL and RED are special cases of the proposed algorithm, when the damping parameter $\alpha=1$. Since the monotone operator is central to our approach, we call it as Monotone Operator Learning (MOL) \cite{pramanik2022improved}. We have introduced theoretical guarantees on the uniqueness of the fixed point of the algorithm, the convergence of the algorithm to the fixed point even when the forward model is low-rank, and the stability of the algorithm to adversarial perturbations. It is also shown that a monotone operator can be constructed as a residual CNN, where a Lipschitz constraint is enforced on the CNN. Traditional DEQ schemes use spectral normalization layer by layer which is a very conservative bound on Lipschitz of the CNN \cite{gilton2021deep} ; our experiments show that the use of this bound translates to poor reconstructions. We had introduced a  Lipschitz regularized training loss (MOL-LR), which can offer a more realistic bound. While the basic theory was introduced in \cite{pramanik2022improved}, the results in \cite{pramanik2022improved} were stated without proofs. In addition, the preliminary experimental results in \cite{pramanik2022improved} were restricted by the small 2D training dataset and the limited number of algorithms we compared against.

The main focus of this work is to rigorously validate the theoretical results in \cite{pramanik2022improved} using experimental data in the context of parallel MRI. We also show the preliminary utility of the algorithm in a 2D+time cardiac cine MRI application, where it is challenging to use unrolled algorithms because of memory constraints. We review the theoretical results \cite{pramanik2022improved} and provide full proofs in the arXiv version \cite{pramanik2022stable}; they are not included in this conference version because of space constraints. In this paper, we compare the proposed  MOL-LR \cite{pramanik2022improved} reconstructions against the DEQ scheme (DE-GRAD) in \cite{gilton2021deep}, MOL-SN, UNET and DU approaches MoDL \cite{aggarwal2018modl}, ADMM-Net \cite{sun2016deep} in the context of 2D parallel MRI. We test our hypothesis to conclude that the performance of the proposed DEQ scheme is comparable to that of the unrolled algorithms, while being significantly more memory efficient. We also evaluate the sensitivity of the algorithms to Gaussian and worst-case input perturbations. This experiment is expected to reveal the benefit of the theoretical bound on robustness in \cite{pramanik2022improved}.

\section{Monotone Operator Learning}
We will now briefly review the MOL algorithm introduced in \cite{pramanik2022improved}. The results in \cite{pramanik2022improved} were stated without proofs. We now restate the main results for accessibility, while a careful reader can find the proofs in the full arXiv version \cite{pramanik2022stable}. 
Consider recovery of an image $\mathbf x \in \mathbb C$ from its noisy under-sampled measurements $\mathbf b$ such that $\mathbf b = \mathcal A(\mathbf x) + \mathbf n$ where $\mathcal A$ is a linear operator and $\mathbf n$ is Gaussian noise. The maximum aposteriori (MAP) estimate of $\mathbf x$ solves for,
\begin{equation}
\label{cs}
C(\mathbf x) = \arg \min_{\mathbf x} \frac{\lambda}{2}\|\mathbf  A \mathbf x - \mathbf b\|_2^2 + \phi(\mathbf x)    
\end{equation}
where $\phi(\mathbf x) = -\log(p(\mathbf x))$ with $p(\mathbf x)$ representing the prior probability density, while $\lambda=\frac{1}{\sigma^2}$ accounts for measurement noise. The minima of \eqref{cs} satisfies the relation:
\begin{equation}
\label{grad}
\underbrace{\lambda \mathbf A^H(\mathbf A \mathbf x - \mathbf b)}_{\mathcal G(\mathbf x)} + \underbrace{\nabla_{\mathbf x} \phi (\mathbf x)}_{\mathcal F(\mathbf x)} = 0.   
\end{equation}
Here, $\mathcal F$ is the gradient of $\phi(\mathbf x)$ and is often termed as the score function. As discussed earlier, one of the strengths of CS formulations is the uniqueness of the solutions. The uniqueness of the solutions that satisfy \eqref{grad} were analyzed in \cite{pramanik2022improved} using the theory of monotone operators, defined as  

\noindent\textbf{Assumption:} The operator $\mathcal F:\mathbb C^M \rightarrow \mathbb C^M$ is m-monotone if:
\begin{equation}
		\label{eqn:monotonicity_definition}
		\Re \left(\Big\langle \mathbf x - \mathbf y, \mathcal F(\mathbf x) - \mathcal F(\mathbf y) \Big\rangle\right) \geq m\|\mathbf x- \mathbf y\|_2^2, \hspace{6pt} m>0 .
\end{equation}
for all $\mathbf x, \mathbf y \in \mathbb C^M$. Here, $\Re(\mathbf x)$ denotes the real part of $\mathbf x$. The following result from \cite{pramanik2022stable} provides the necessary and sufficient condition for the solution of \eqref{grad} to be unique.
\begin{prop}\cite{pramanik2022stable}
	\label{propv4}
	The fixed point of \eqref{grad} is unique for a specific $\mathbf b$, iff $\mathcal F$ is $m$-monotone with $m>0$.
\end{prop}
\begin{prop}\cite{pramanik2022stable}
	\label{resnet}
	$\mathcal F:\mathbb C^M \rightarrow \mathbb C^M$ is $m$-monotone if it can be represented as a residual CNN, $\mathcal F=\mathcal I-\mathcal H_{\theta}$, where $\mathcal H_{\theta}:\mathbb C^M \rightarrow \mathbb C^M$ has a Lipschitz constant, $L[\mathcal H_{\theta}] = (1-m); ~~0<m<1$. 
\end{prop}
\subsection{Proposed MOL algorithm}
With these results, we use a forward-backward splitting of \eqref{grad} for $\alpha > 0$, to yield the iterative algorithm $\mathbf x_{n+1} = (I + \alpha \mathcal G)^{-1}(I - \alpha \mathcal F)(\mathbf x_n)$, which is expanded as
\begin{eqnarray}\nonumber
	\label{fp}
	\mathbf x_{n+1} & = &  \underbrace{(\mathcal I + \alpha \lambda \mathbf A^H \mathbf A)^{-1}\Big((1-\alpha)\mathbf x_n + \alpha \mathcal H_{\theta}(\mathbf x_n) \Big)}_{\mathcal T_{\rm MOL}(\mathbf x_n)} +\\
	&&\qquad \underbrace{(\mathcal I + \alpha \lambda \mathbf A^H \mathbf A)^{-1}\Big( \alpha \lambda \mathbf A^H \mathbf b\Big)}_{\mathbf z}
\end{eqnarray}
\begin{prop}
	\label{thmiv3}
	Consider the algorithm specified by \eqref{fp}, where $\mathcal F$ is an $m$-monotone operator. Assume that \eqref{fp} has a fixed point specified by $\mathbf x^*(\mathbf b)$. Then, convergence is guaranteed for an arbitrary $\mathbf A$ operator when 
	\begin{equation}
		\label{condition}
		\alpha  < \frac{2 m}{(2-m)^2} = \alpha_{\rm max}
	\end{equation}
\end{prop}
The algorithms \cite{aggarwal2018modl,hammernik2018learning,romano2017little}
correspond to the special case of $\alpha = 1$. Setting $\alpha=1$ in \eqref{condition}, we see that the algorithm will converge if $m \geq 3-\sqrt{5}=0.76$
or $L[\mathcal H_{\theta}]\leq0.24$. The denoising ability of a network is dependent on its Lipschitz bound; a smaller $L[\mathcal H_{\theta}]$ bound translates to poor performance of the resulting MOL algorithm. The use of the damping factor $\alpha <1$ allows us to use denoising networks $\mathcal H_{\theta}$ with larger Lipschitz bounds and hence improved denoising performance. For instance, if we choose $m=0.1; L[\mathcal H_{\theta}]=0.9$, from \eqref{condition}, the algorithm will converge if $\alpha < 0.055$.

The following result bounds the perturbation in the solutions in response to worst case measurement noise in $\mathbf b$.
\begin{prop}\cite{pramanik2022improved}, 
	\label{robustness}
	Consider $\mathbf z_1$ and $\mathbf z_2$ to be measurements with $\boldsymbol\delta =\mathbf z_2-\mathbf z_1$ as the perturbation. Let the corresponding outputs of the MOL algorithm be $\mathbf x^*(\mathbf z_1)$ and $\mathbf x^*(\mathbf z_2)$, respectively, with $\Delta= \mathbf x^*(\mathbf z_2)-\mathbf x^*(\mathbf z_1)$ as the output perturbation, $\|\Delta\|_2
		\leq\frac{\alpha \lambda }{1-\sqrt{1 - 2\alpha m + \alpha^2(2 - m)^2}}~\|\boldsymbol \delta\|_2.$
When $\alpha$ is small, we have $\lim_{\alpha\rightarrow 0}\|\Delta\|_2 \leq \frac{\lambda}{m} \|\boldsymbol\delta\|_2$.
\end{prop}

We note from Proposition 2.2 that a monotone $\mathcal F = \mathcal I - \mathcal H_{\theta}$ can be learned by constraining the Lipschitz constant of $\mathcal H_{\theta}$. A common approach is spectral normalization \cite{gilton2021deep,ryu2019plug}. However, this is a conservative estimate, often translating to lower performance.
We use the Lipschitz estimate \cite{bungert2021clip}, $l\left[\mathcal H_{\theta}\right] = \max_{\mathbf x \in S}\overbrace{\sup_{\boldsymbol\eta} \frac{\|\mathcal H_{\theta}(\mathbf x + \boldsymbol\eta) -\mathcal H_{\theta}(\mathbf x)\|_2^2}{\|\boldsymbol\eta\|_2^2}}^{P(\mathbf x)}$.
This estimate is less conservative than the one using spectral normalization. Our experiments show that the use of this estimate can indeed result in algorithms with convergence and robustness as predicted by the theory. In the supervised learning setting, we propose to minimize, $V(\theta) = \sum_{i=0}^{N_t} \|\mathbf x_i^* - \mathbf x_{i}\|_2^2 ~\mbox{s.t.}~ P\left(\mathbf x_i^*\right)\leq T; i=0,..,N_t.$ 
	Here, $T = 1-m$. The above loss function is minimized with respect to parameters $\theta$ of the CNN $\mathcal H_{\theta}$ using a log-barrier approach, 
 \begin{equation}
 \theta^* = \arg \min_{\theta}~\underbrace{\sum_{i=0}^{N_t} \Big(\|\mathbf x_i^* - \mathbf x_{i}\|_2^2 -  \beta \log \left(T - P\left(\mathbf x_i^*\right)\right)\Big)}_{C_i}.
\end{equation}
 $\mathbf x_i^*$ is a fixed point of \eqref{fp} that is dependent on the CNN parameters $\theta$. $\mathbf x_i; i=0,..,N_t$ and $\mathbf b_i, i=0,..,N_t$ are the ground truth images in the training dataset and their undersampled measurements, respectively.

\section{Experiments and Results}
The experiments in this paper build upon the preliminary results in \cite{pramanik2022improved}, where the proposed MOL-LR (Lipschitz regularized) algorithm was compared against MOL-SN (spectral normalization) and MoDL in terms of performance. However, the models were trained with few datasets (90 slices from each of the 4 training subjects). In addition, the experiments in \cite{pramanik2022improved} did not study the robustness of the proposed algorithm shown in Proposition 2.4 or the benefit of the scheme in high-dimensional applications, in comparison to other unrolled methods. In this work, we address above limitations of \cite{pramanik2022improved} by performing experiments on a large publicly available brain MRI data from the Calgary Campinas Dataset (CCP) \cite{souza2018open}. CCP consists of 12-channel (coil) T1-weighted 2D brain data from 117 subjects. We split data from 40, 7 and 20 subjects into training, validation and testing sets respectively. Cartesian 2D non-uniform variable density masks are used for 4-fold undersampling of the datasets.

\begin{figure*}[!ht]
	\centering
	\includegraphics[scale=0.8,keepaspectratio=true,trim={3cm 9cm 3cm 9.7cm},clip]{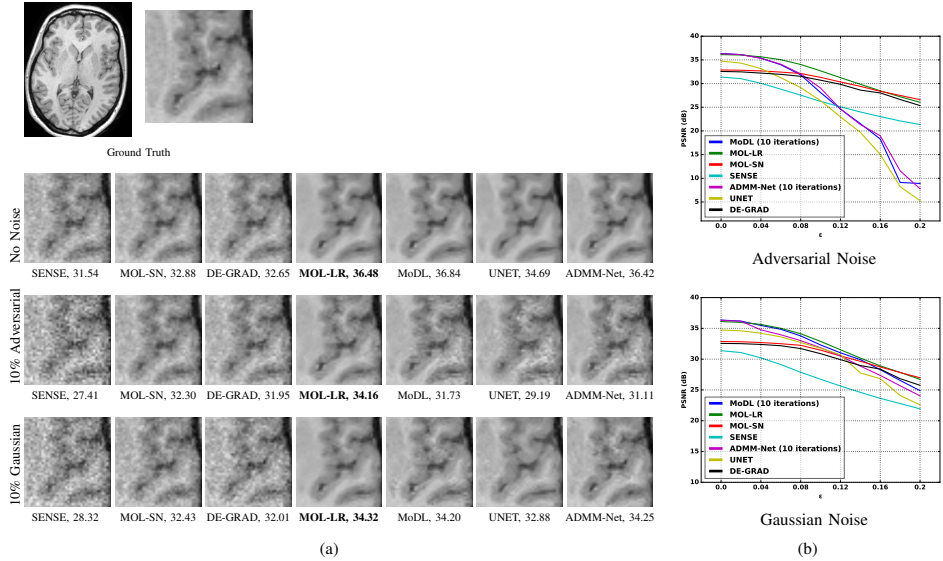}
	\caption{Sensitivity of the algorithms to input perturbations. In (a), the rows correspond to reconstructed magnitude results of 4x accelerated multi-channel brain data for no noise, 10\% worst-case (adversarial) and 10\% gaussian noise, respectively. The PSNR (dB) values are reported for each case. The data was undersampled using a Cartesian 2D non-uniform variable-density mask; (b) shows plots of PSNR vs percentage of Adversarial or Gaussian Noise in terms of $\epsilon$ (from 0\% to 20\%).
}
	\label{fig:brain_rob_compare}
\end{figure*}

We compare the performance of MOL-LR and its robustness to against DEQ methods (DE-GRAD, MOL-SN), UNET and DUs (MoDL, ADMM-Net) in the first row of images in Fig. \ref{fig:brain_rob_compare} for recovery from 4-fold undersampled data. The SENSE, MOL-SN and DE-GRAD reconstructions are relatively more noisy compared to other methods. SENSE is a CS algorithm, utilizing coil sensitivities obtained from calibration scans. It fails to perform at such a high acceleration factor. MOL-SN and DE-GRAD are DEQs with Lipschitz of the network bounded by spectral normalization of each layer. Since, this is a very strict bound, the CNNs attain a much lower Lipschitz compared to MOL-LR, which translates to poor performance. MOL-LR performs at par with DUs (10-iterations of MoDL, ADMM-Net) and outperforms UNET.

We study degradation of the reconstructions, when the input measurements are corrupted by Gaussian or worst-case perturbations determined using an adversarial attack. The experiments in the third row of Fig. \ref{fig:brain_rob_compare}(a) show that all the methods perform similarly on Gaussian noise. By contrast, the experiments in the second row of Fig. \ref{fig:brain_rob_compare}(a) show that MOL-SN and MOL-LR are relatively more robust to adversarial noise, compared to MoDL, UNET, and ADMM-Net. Both MOL-SN and MOL-LR are associated with guarantees on robustness. The plots in Fig. \ref{fig:brain_rob_compare}(b) shows drop in PSNR with respect to percentage of Adversarial/Gaussian noise (0\% to 20\% for $\epsilon = [0.0, 0.2]$) and shows similar trend. 

\begin{figure}[h!]%

    \centering
	\includegraphics[width=0.4\textwidth,keepaspectratio=true,trim={1.5cm 8.6cm 1.6cm 9cm},clip]{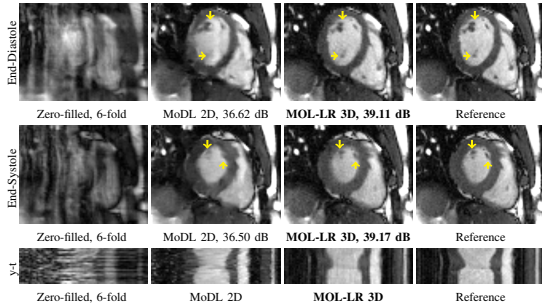}
    
    \caption{MOL recovery of 2D+time cine data at 6x acceleration. PSNR (dB) values are reported for each case. The data is retrospectively undersampled using a Poisson density sampling pattern.}%
    \label{fig:cardiac}%
\end{figure}

We demonstrate the benefit of reduced memory demand in MOL over unrolled networks through experiments on 2D+time cardiac data, run on a 16GB GPU. While MOL-LR 3D performs recovery in 3D, MoDL \cite{aggarwal2018modl} could do it only in 2D due to lack of memory for multiple unrolls. MOL-LR 3D outperforms MoDL 2D as shown in Fig. \ref{fig:cardiac}. MoDL 2D's poor performance can be attributed to lack of temporal information which is well utilized by MOL-LR 3D.

\section{Conclusion}
In this paper, we demonstrate the benefits of a DEQ-based monotone operator learning (MOL), proposed in \cite{pramanik2022improved}. Similar to compressed sensing algorithms with convex priors, MOL possesses guarantees for uniqueness of the solution, convergence to the fixed-point and stability to input perturbations. MOL is found to have ten-fold reduction in memory consumption compared to one consumed by 10-iterations of unrolled networks. Results show that MOL is significantly more robust to adversarial attacks and performs at par on noiseless data, compared to unrolled networks. We apply MOL to higher-dimensional problems which cannot be done for unrolled networks due to memory constraints.

\section{Compliance with Ethical Standards}
This research study was conducted using human subject data. Approval was granted by the Ethics Committee of the institute where the data were acquired.

\bibliographystyle{IEEEbib}
\bibliography{refs}
\end{document}